\documentclass[letterpaper, 10 pt, conference]{ieeeconf}  

\IEEEoverridecommandlockouts                              

\usepackage{graphicx} 
\usepackage{amsmath} 
\usepackage{multirow, booktabs}
\usepackage[T1]{fontenc}
\usepackage{cite}
\usepackage{algorithm}
\usepackage{algorithmic}

\usepackage{amsfonts}
\usepackage{hyperref}

\newcommand{\R}{\mathbb{R}}

\newcommand{\EE}{\mathbb{E}}

\newcommand{\sset}{\mathcal{S}}
\newcommand{\aset}{\mathcal{A}}

\newcommand{\trans}{\mathcal{P}}

\title{\LARGE \bf
Cooperative-Competitive Team Play of Real-World Craft Robots
}

\author{Rui Zhao$^{1*}$, Xihui Li$^{1,2*}$, Yizheng Zhang$^{1*}$, Yuzhen Liu$^{1*}$, \\
    Zhong Zhang$^{1}$, Yufeng Zhang$^{1}$, Cheng Zhou$^{1}$, Zhengyou Zhang$^{1}$, Lei Han$^{1}$
\thanks{
$^{*}$Equal contribution.
$^{1}$The authors are with Tencent Robotics X Laboratory, Shenzhen, Guangdong, China.
$^{2}$The author is with Tsinghua University, Shenzhen, Guangdong, China.
{Corresponding to \tt\small rui.zhao.ml@gmail.com, leihan.cs@gmail.com}}
}

\begin{document}

\maketitle
\thispagestyle{empty}
\pagestyle{empty}

\begin{abstract}
Multi-agent deep Reinforcement Learning (RL) has made significant progress in developing intelligent game-playing agents in recent years. However, the efficient training of collective robots using multi-agent RL and the transfer of learned policies to real-world applications remain open research questions. In this work, we first develop a comprehensive robotic system, including simulation, distributed learning framework, and physical robot components. We then propose and evaluate reinforcement learning techniques designed for efficient training of cooperative and competitive policies on this platform. To address the challenges of multi-agent sim-to-real transfer, we introduce Out of Distribution State Initialization (OODSI) to mitigate the impact of the sim-to-real gap. In the experiments, OODSI improves the Sim2Real performance by 20\%. We demonstrate the effectiveness of our approach through experiments with a multi-robot car competitive game and a cooperative task in real-world settings.
\end{abstract}

\section{INTRODUCTION}

Deep Reinforcement Learning (RL) has demonstrated superhuman performance in multi-player games, such as Dota~\cite{berner2019dota}, StarCraft~\cite{vinyals2019grandmaster,han2020tstarbot}, and Hide and Seek~\cite{baker2019emergent}. 
In robotics, RL has been applied to different kinds of robots. 
For single-robot systems with high-dimensional nonlinear dynamics, RL has advanced locomotion behaviors of real-world quadrupedal robots~\cite{hwangbo2019learning,lee2020learning,miki2022learning,margolis2022rapid,chen2023learning,feng2023genloco,fu2023deep,li2023learning,han2023lifelike,hoeller2023anymal,zhang2024learning} and bipedal robots~\cite{kumar2022adapting,radosavovic2023learning,haarnoja2023learning,li2024reinforcement}, empowered these robots to learn agile motions.
For multi-robot scenarios, multi-agent RL has shown progress in learning cooperative robotic systems, including mobile robots~\cite{touzet2004distributed,tuci2018cooperative,goldhoorn2018searching,mitchell2019multi,koh2020cooperative,candela2022transferring,torbati2023marbler,werner2023dynamic}, drones~\cite{shi2022marl,yang2023partially}, and robotic arms and hands~\cite{beaussant2021delay,huang2023dynamic}. 
Multi-agent cooperation can greatly improve the efficiency of transportation~\cite{park2023multi}, warehouse management~\cite{wurman2008coordinating}, and construction~\cite{ren2004multi}, which further potentially promotes social production and development.
Despite the agility behaviors learned by robots with deep RL, the ability to cooperate with others, i.e., collective intelligence, is of great importance for multi-agent robotic systems.

The cooperative effort of a group allows for greater achievements than any individual could achieve.
In the natural world, ants evolve collective intelligence through millions of years. To maximize the interest of the species, ants divide labors and assign sub-tasks for different roles. Through teamwork, ants achieve great results, such as building complex underground nets~\cite{alexander1974evolution}.

Enabling robots to emergent similar swarm intelligence has been a long-standing challenge. 
As the complexity and the number of robots increases, manually building a multi-agent control system~\cite{wurman2008coordinating} becomes impractical.
Unlike conventional methods that depend on manual design or in-depth knowledge of system dynamics, RL can potentially deliver more robust policies since it needs minimal domain expertise, only a reward signal.
Recent advances in deep RL, such as self-play~\cite{silver2017mastering}, could potentially transcend multi-agent intelligence from games into robotics.

In contrast to RL in game AI, training cooperative policies for robots poses different challenges. 
Video games are natural environments in which to train AI systems. 
There is no difference between the training environments and the testing environments. 
In robotics, collecting real-world data from physical robots is extremely expensive and time-consuming. 
To mitigate this problem, we need to build representative and efficient simulations to generate data in scale for training data-driven approaches. 
How to balance the complexity and the fidelity of the simulation is one of the challenges. 
Practical design choices are necessary for an efficient learning-based robotic system. 
It is impossible to build a simulated environment that is exactly the same as the real world. 
The Sim-to-Real (Sim2Real) gap greatly affects the performance of multi-agent reinforcement learning for robotics, which is a major challenge.

Previous works on cooperative-competitive multi-agent RL via self-play are mainly in the virtual world, such as StarCraft~\cite{samvelyan2019starcraft}, MineCraft~\cite{perez2019multi}, and simulated football~\cite{liu2022motor}. 
Examples of using RL for multi-robot cooperation and multi-team competition in the real world remain elusive.
This is partly attributed to the increased complexity of interactions among agents and teams.
This complexity challenges the reliance of modern RL algorithms on large amounts of real-world data. 
Recent works have bypassed real-world data collection by training policies in simulations via domain randomization~\cite{sadeghi2016cad2rl,tobin2017domain,matas2018sim,peng2018sim,andrychowicz2020learning,peng2020learning}. 
However, domain randomization has scalability issues.
As the task becomes more complex, its simulated environment must utilize more randomizations. 
For example, a single-agent task may only need to randomize properties associated with the agent, but if multiple agents are introduced, the relative properties among agents need to be randomized as well. 
Parallel to the scalability challenges and the hyperparameter tunning of the randomizations, if too many randomizations are used, the learned policy might be too conservative~\cite{ramos2019bayessim}. 
With these conservative policies, the agents might be unable to properly solve the original task.
Compared to single-robot systems, multi-robot systems pose more challenges in efficient learning and sim-to-real transfer, as the dimensionality and complexity increase.
More practical and efficient sim-to-real techniques are needed for the multi-agent settings.

We propose to utilize guided RL~\cite{esser2022guided} to increase learning efficiency and develop a novel sim-to-real method named Out of Distribution State Initialization (OODSI) to mitigate the discrepancy of multi-agent dynamics between simulation and the real world. 
To test the cooperative-competitive policy and its resilience against the sim-to-real gap, we build a multi-team game environment with craft robots in the real world. 
The craft-robots are essentially mobile robots with the ability to move and build constructions with different objects, such as blocks and slopes. 
The robots can move and unfold the slopes to build different constructions and change the layout of the landscape.
We also build the corresponding simulations, the first simulation is based on pyBullet, which has discrete observation and action spaces and is fast to run rollouts for training. The second simulation is based on Gazebo, which has continuous signals and is relatively slow but more realistic to the physical robot scenarios. 
We use the pyBullet-based simulation for training. 
We use the Gazebo-based simulation for testing before deploying the policies on the real robots.

In this robotic platform, we design two games. The first one is a two-team competition game. In this competition, there are limited amount of resources, including the block and slopes. The goal for each team is to build a second-floor structure. The resources is only enough for one team to finish the building. Therefore the two teams need to compete with each other to secure objects and build fast. During learning, we observe the evolution of attack and defense strategies. 
The second game is a simple word-building task, which is used for more comprehensive ablation tests and in-depth analysis of the algorithms. 

In this paper, we build a complete robotic AI system for multi-agent mobile robots, including the simulations, learning framework, and physical robots. We show that design choices are important to make the training process more efficient. We find that guided reinforcement learning~\cite{esser2022guided} can speed up learning and propose a technique to inject human guidance into reinforcement learning. 
After training the policies in simulation, we deploy the policies directly on the real robots, and have relative low success rate. 
We propose the OODSI method to train robust policies to tackle the sim-to-real gap. 
The main contribution of this paper is two-fold: 
first, we demonstrate how to use multi-agent RL to train real-world mobile robots to develop competitive and cooperative team strategies through self-play. 
Secondly, due to the Sim2Real gap of asynchronous actions of multi-robots, the learned cooperation strategies are infeasible during deployment. We propose OODSI method to combat the Sim2Real gap in the multi-agent RL settings.

\section{Robot Craft Arena}
\label{sec:robot_craft_arena}
In the video game MineCraft, players can build various customized constructions using the unit blocks. 
Inspired by this, we build a real-world robot craft arena, which is essentially a robotic construction platform~\cite{xu2022reccraft,zhao2023craftenv}. 
In this arena, the robot can interact with different objects and build constructions.
The goal of the agents is to build a target construction as a team, or as two teams to compete with each other.
The robot craft arena consists of three elements, i.e., the craft robots, blocks, and slopes. 
The craft-robots are mobile manipulation vehicles, which can transport the blocks and slopes.
The robot utilizes a front camera and ApirlTags for localization.
The block is a hollow cube. 
The craft robots can move underneath the block, lift it, move it, and place it at any accessible location in the arena. 
The slope is a folding ramp block. 
With the slope as the building material, the craft robots can move to higher floors of the construction. 

Based on the real-world specifications, we build a simulation using pyBullet for training. 
In order to train the RL agents efficiently, we discretize the observation space and the action space. 
The agent perceives a list of attributes as observation, which describes the position, orientation, and categorical information of all the objects in the arena.   
After processing the observation, the agent can choose one out of eleven actions, including moving forward, back, left, or right, turning left or right, lifting the object, i.e., block or slope, dropping the object, folding or unfolding the slope, or stop i.e., no-operation. 

\subsection{Real-World Robot Constrains}
\label{sec:real-world-constrains}
In the simulation, we consider the real-world constraints of the robots. 
For example, when the robot wants to move into the slope and move it, it needs to first adjust its orientation to make the camera face outwards and then roll back into the slope. This is to avoid losing track of AprilTags, as the slope has an opening only on one side. 
Similarly, in the case of a block, when the robot is underneath it and wants to move out, it needs to first turn to face forward and then move out of the block. It cannot move directly left or right to exit the block.
There are other similar constraints. 
To avoid actions that do not comply with constraints, we utilize action masks to block unfeasible actions and replace them with stop actions.
However, with action masks, we can avoid unfeasible and dangerous actions. 
However, these real-world constraints make the task more difficult to solve. 
For instance, to find the solution to entering the slope, the robot needs to first explore the orientation of its rear part facing the entrance of the slope and then move back in. Given an initial random policy, the product probability of this consecutive action sequence is relatively low, which means it is difficult for the agent to explore the solution. 
These real-world robot constraints make the training process challenging.

\subsection{The Multi-Agent Sim-to-Real Gap}
\label{sec:sim-to-real-gap}
Multi-robot cooperation and competition pose unique sim-to-sim and sim-to-real challenges. 
In the training simulation, the observation and action space are discrete. 
For each time step, given the current state, multiple agents choose valid actions from the action space. 
The simulation takes in all the actions and produces the next state all at once. 
The action execution of multiple agents in simulation is synchronized.
However, in reality, in different situations, agents' actions take different amounts of time. 
\begin{algorithm}
    \caption{Out of Distribution State Initialization (OODSI)}
    \label{algo:oodsi}
    \begin{algorithmic}[1]
    \STATE Train policy $\pi_{\theta}$ in the training simulation (pyBullet)
    \STATE Use $\pi_{\theta}$ to interact with the more realistic simulation (Gazebo) or the real-world environment to collect trajectories $\mathcal{T}$, which contains OOD states $s^{OOD}$ 
    \STATE Sample initial states $s_{0} \sim \mathcal{T}$ for training
    \STATE Retrain policy $\pi_{\theta}$ in the training simulation
    \end{algorithmic}
\end{algorithm}
For example, if the robot car starts from standing still and moves forward by one block distance, it might take one second. 
If the robot car is moving forward and keeps moving forward by one block, it takes less than one second. 
Compared to consecutively moving forward by a number of blocks, the action sequence of the same length consisting of moving forward and turning takes a longer time. 
As a consequence, in the real world, for a time step, some of the agents finished the commanded actions and others could not. 
For the robots that have not finished the actions, they will first finish the current action and then take in a new action.
This causes a unique sim-to-real gap in the setting of multi-robot cooperation and competition. 
The dynamics have changed. For the same starting state, after executing the same agents' actions, the agents might perceive different next states due to action delays and the asynchronization of multi-agent actions.
In reality, the action execution process is asynchronous, while in the training simulation, it is synchronous.
This sim-to-real gap makes the deployment of simulation-trained multi-agent policy difficult to transfer to the real world.
To reproduce this sim-to-real gap in simulation for quick testing and evaluation, we build a Gazebo-based simulation, which has continuous action space as in the real world. The Gazebo-based simulation also utilizes the same asynchronous action execution process as the real-world robots. Although the Gazebo-based simulation is more realistic it is significantly slower compared to the discrete pyBullet-based simulation, which makes the Gazebo-based simulation inadequate for training. We only use the Gazebo-based simulation to test the sim-to-real gap before deploying the policies on the real-world robots. For simplicity, we refer to the sim-to-sim gap between the pyBullet-based simulation and the Gazebo-based simulation as the sim-to-real gap in the rest of the paper.

\section{Preliminaries}
A Markov decision process (MDP) is represented by a tuple $M = (\sset, \aset, \trans, r, \rho_0, T)$, where $\sset$ is a state set, $\aset$ an action set, $\trans: \sset \times \aset \times \sset \rightarrow \mathbb{R}_{+}$ is a transition probability distribution, $r: \sset \times \aset \rightarrow \R$ is a reward function, $\rho_0: \sset \to \mathbb{R}_+$ is a start state distribution, and $T$ is the horizon. 
For multi-agent settings, the transition probability distribution and the reward function are depend on the actions from all agents, i.e.,  $\trans: \sset \times \aset^1 \times \aset^2\ ... \times \aset^{N} \times \sset \rightarrow \mathbb{R}_{+}$ and $r: \sset \times \aset^1 \times \aset^2\ ... \times \aset^{N} \rightarrow \R$, where $i \in \{1,2,...,N\}$ is the index for the $i$-th agent.
For the multi-team settings, the definition of the reward function implies the game is cooperative or competitive. 
For example, in a two-team zero-sum game, if we define $r^1 + r^2 = 0$, then the game is competitive, as one team wins and the other loses. 
The goal of RL is to learn a policy $\pi_{\theta}: \sset \times \aset \to \mathbb{R}_+$ parametrized by $\theta$ that maximizes the expected accumulated reward $ R(\pi, s_0) := \EE_{\tau|s_0}[ \sum_{t=0}^T r(s_t, a_t) ]$, i.e., return, $ \eta_{\rho_0}(\pi_\theta) = \EE_{s_0\sim\rho_0} R(\pi, s_0)$, starting from a $s_0\sim \rho_0$, where $\tau = (s_0, a_0, , \ldots, a_{T-1}, s_T)$ denotes a complete trajectory, with $a_t \sim \pi_\theta(a_t|s_t)$, and $s_{t+1} \sim \trans(s_{t+1} | s_t, a_t)$. 
Policy optimization methods, such as PPO~\cite{schulman2017proximal}, iteratively collect trajectories and then use them to improve the current policy~\cite{sutton2018reinforcement}. 
In practice, action masking is utilized as a common technique to avoid generating invalid actions in discrete action spaces~\cite{vinyals2017starcraft,berner2019dota,ye2020mastering,sun2020tleague,han2020tstarbot}.

\subsection{Guided RL}
In real-world robotics applications, RL still faces challenges. 
Guiding the learning process with additional knowledge offers potential solutions, which leverage the data-driven approaches and knowledge-based approaches.
The goal of guided RL is to accelerate learning efficiency and improve the success rate for real-world robotics deployment via the integration of additional knowledge into the learning process~\cite{esser2022guided}.
Recent advances in RL have testified the advantages of utilizing prior knowledge, such as natural languages~\cite{kaplan2017beating,ahn2022can} and programs~\cite{sun2019program,chang2022mapp} to accelerate policy learning. 

\begin{figure}
    \centering
    \includegraphics[width=0.45\textwidth]{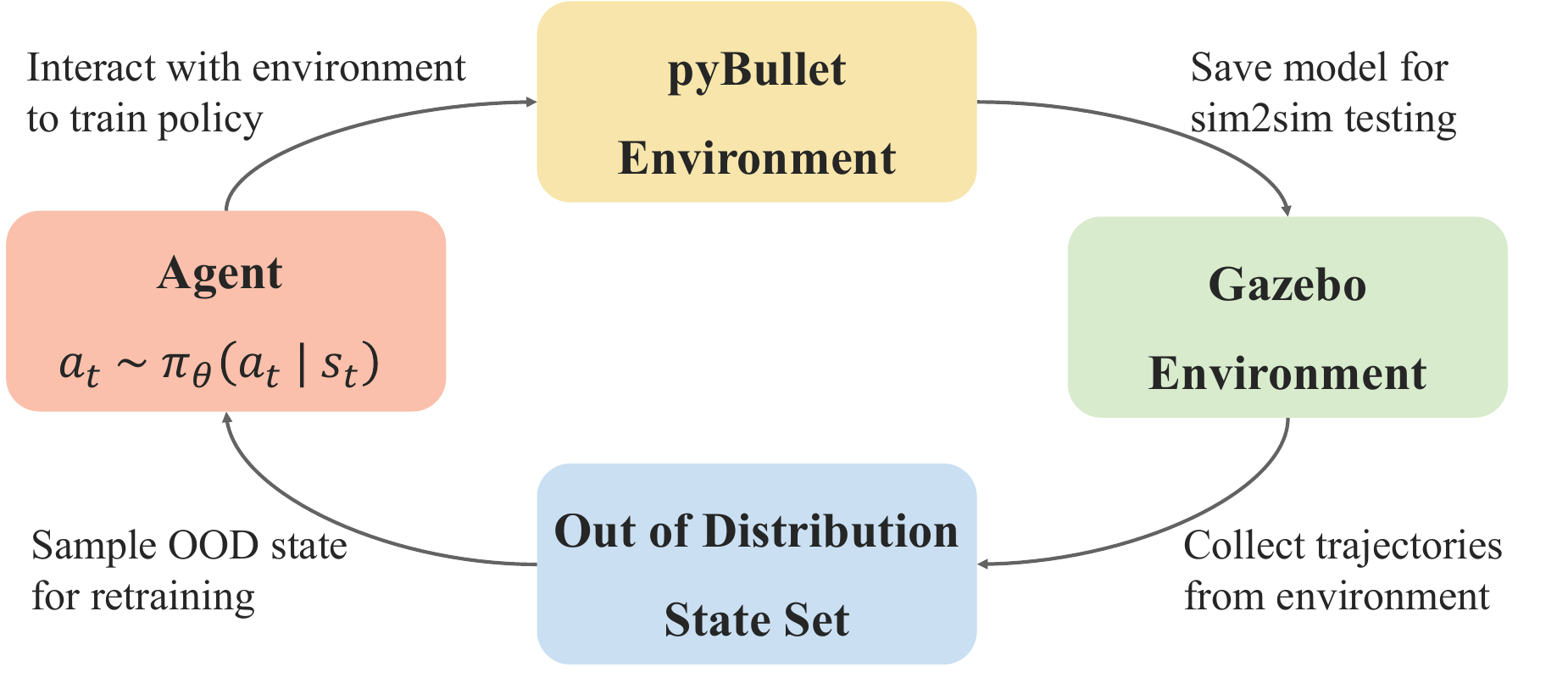}
    \caption{\textbf{Out of Distribution State Initialization }}
    \label{fig:oodsi}
\end{figure}

\subsection{Modifying Start State Distribution in RL}
In the literature of RL, researchers directly modify the start state distribution $\rho_0$ to accelerate learning in an MDP. 
Popov et al.~\cite{popov2017data} propose to modify the start state distribution $\rho_0$ to be uniform among states visited by expert demonstrations to speed up learning.
Florensa et al.~\cite{florensa2017reverse} utilize a set of start states increasingly far from the goal to gradually train robots to conduct navigation and manipulation tasks in simulation.
In our work, we propose a method based on modifying the start state distribution $\rho_0$ to make the policies more robust against the sim-to-real gap.

\section{METHODOLOGY}
In this section, we propose two innovations to tackle the aforementioned two challenges, respectively, i.e., efficient learning under real-world robot constraints and the multi-agent sim-to-real gap. 
The first innovation is guided RL via action masking, which can be considered a practical technique.
The second one is a method named Out of Distribution State Initialization (OODSI) for sim-to-real transfer.

\begin{figure*}
  \centering
  \begin{minipage}[b]{0.37\textwidth}
    \centering
    \includegraphics[width=\textwidth]{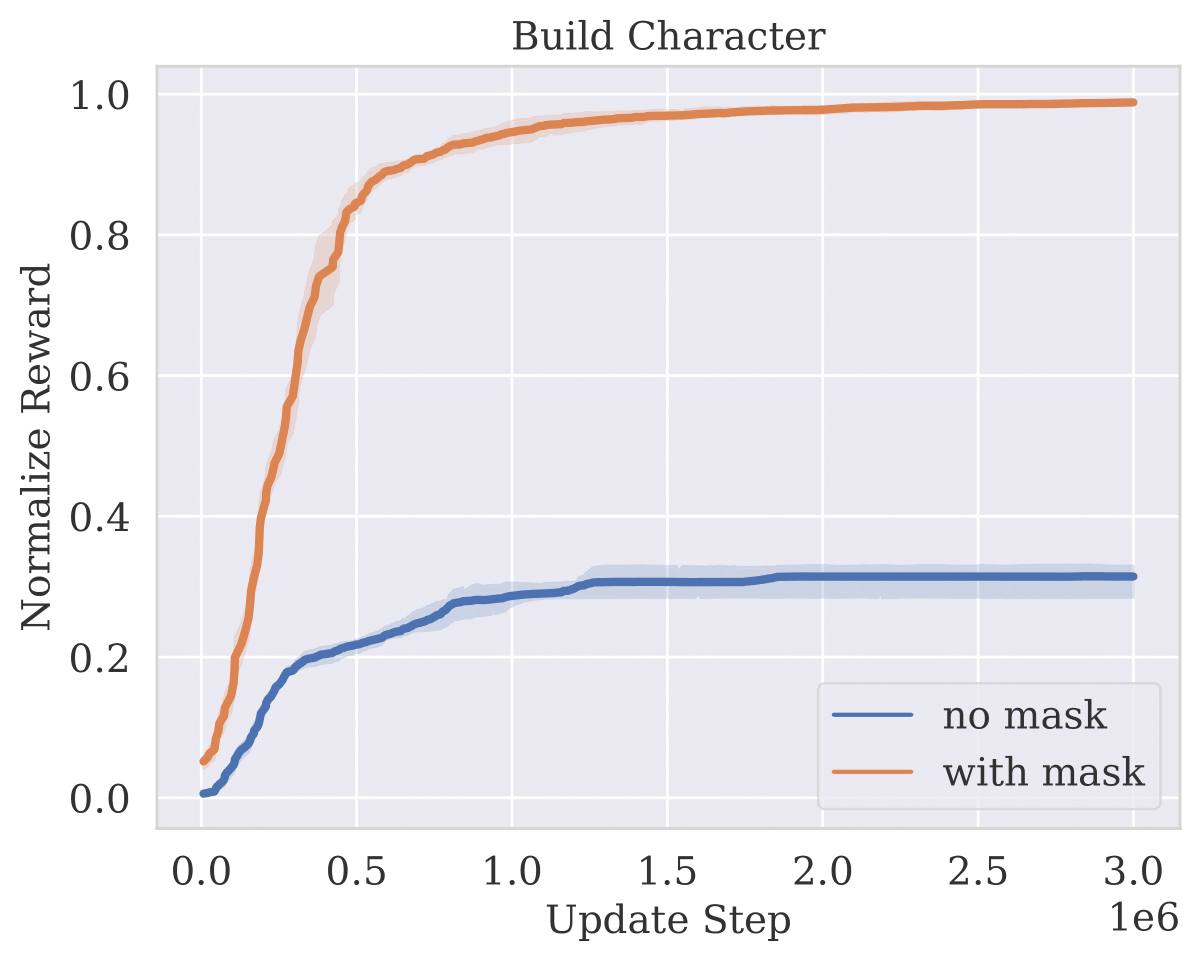}
    \caption{Training curves w/o action masking}
    \label{fig:action_mask_collaboration}
  \end{minipage}
  \hspace{1.0cm}
  \begin{minipage}[b]{0.37\textwidth}
    \centering
    \includegraphics[width=\textwidth]{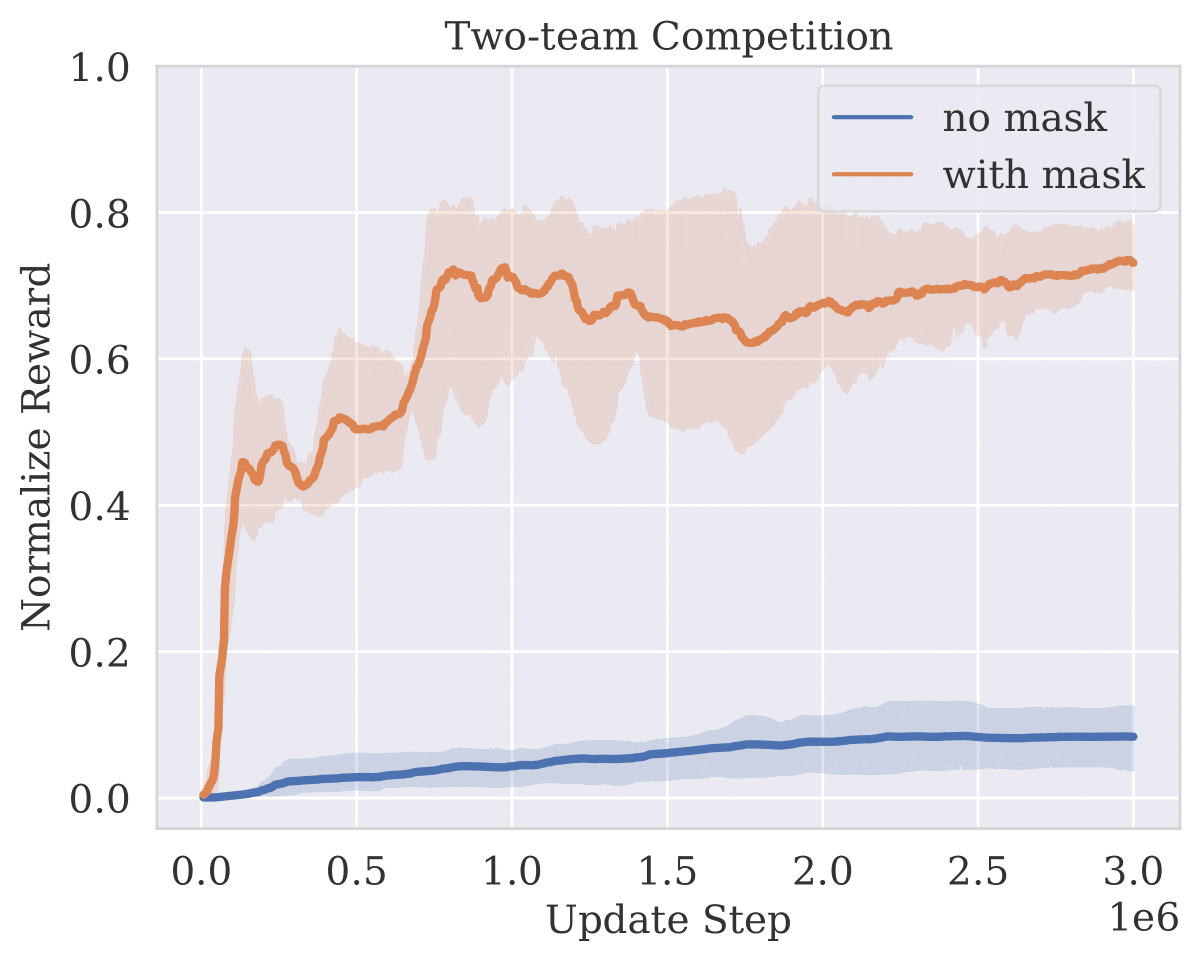}
    \caption{Training curves w/o action masking}
    \label{fig:action_mask_competition}
  \end{minipage}
\end{figure*}

\subsection{Guided RL via Action Masking}
\label{sec:action_mask}
In the robot craft arena, there are real-world constraints, see Section~\ref{sec:real-world-constrains}, such as the specific orientation for the robot to move into the slope or move out of the block.
We implemented rule-based sub-routine planners through the action masks to guide the robot to efficiently explore the environment, such as moving into the slope, etc.
To introduce the guidance, given a state, along with the set of invalid actions, we further block the actions that are not beneficial for the downstream task. 
This reduces the exploration space for the agents.
The implemented rule-based sub-routine planners can also be seen as parts of the environment. 
Similar to the game environments, the agent does not need to consider low-level details, when performing actions, such as picking up a box in MineCraft or walking toward the enemies in StarCraft.
This design choice simplifies the task for the RL agents.
The RL agent can focus on learning high-level strategies. 
The combination of rule-based sub-routine local planners and the data-driven RL policies enable the agents to efficiently learn complex real-world robotics tasks.

\subsection{Out of Distribution State Initialization}

We propose Out of Distribution State Initialization (OODSI) to tackle the multi-agent sim-to-real gap.
As mentioned in Section~\ref{sec:sim-to-real-gap}, in the training environment, i.e., the pyBullet-based environment, the action execution of multiple agents is synchronous and the actions are discrete. 
Whereas in the testing environment, i.e., the Gazebo-based environment and the real-world robot craft arena, the action execution is asynchronous and continuous.  
This discrepancy between the training and the testing environment leads to the sim-to-real gap.
Under the sim-to-real gap, the dynamics are different between the training environment and the testing environment. 
To be more specific, under the same current state, $s_t$, executing the same multi-agent actions, $a_1, a_2,\ ... a_n$, the next states in the training environment, $s_{t+1}^{train}$, could be different as the next states in the testing environments, $s_{t+1}^{test}$.
With respect to the trained agents, these unseen states from the testing environments are Out of Distribution (OOD) states $s^{OOD}$.
We propose to add these OOD states $s^{OOD}$ collected from the testing environments to the start state distribution, $\rho_0$, in the training process to train the agent to learn robust policies against OOD states.
After training with the OOD states, when encountering the OOD states, the agents would still know how to handle the situation and complete the task. 
Our approach can be understood as sequentially composing locally stabilizing policies by growing trajectories from an OOD state to the next OOD state, of which the idea bears similarity with the work of Tedrake et al.~\cite{tedrake2010lqr} and the work of Burridge et al.~\cite{burridge1999sequential}.
We summarize OODSI method in Algorithm~\ref{algo:oodsi} and Figure~\ref{fig:oodsi}.

\subsection{States, Actions and Rewards}
We construct the observation of the agent's current position and the positions of other agents. 
The actions of the craft robots include moving in horizontal and vertical directions and interacting with different objects in the environment. 
The craft robot is designed to possess the ability to interact with objects in the environment in various ways, such as lifting, moving, and unloading a block or a slope, and folding and unfolding a slope. 
In the craft arena, we define two tasks. 
The first one is a cooperative building characters task. 
The second one is a competitive building two-floor structure task between two teams.
For the cooperative task, we first define a target building, which consists of multiple blocks and slopes. 
When the agents build one more block right, the agents receive a positive reward $r_{build}$.
After the agents successfully complete the building, the agents receive a completion reward $r_{completion}$, which is designed to be larger than the building reward $r_{build}$ to encourage the agents to finish the whole building instead of leaving several pieces behind. 
In the case of the two-team building competition, the resources are limited. 
There is a block missing, hence only one of the two teams can finish their building and win the game receiving $r_{completion}$. 
And the other team loses in the game.
The two-team competition is a zero-sum game.

\subsection{Learning Framework}

\begin{table*}[]
    \caption{Success rate in pyBullet and Gazebo}
    \label{tab:sim2real}
    \renewcommand{\arraystretch}{1.3}
    \centering
    \begin{tabular}{llcccc}
    \toprule
    \multicolumn{2}{l}{}       & \multicolumn{2}{c}{Build Character} & \multicolumn{2}{c}{Two-team Competition} \\ \cmidrule(r){3-4} \cmidrule(r){5-6}
    \multicolumn{2}{l}{Method}       & pyBullet         & Gazebo      & pyBullet         & Gazebo       \\  \cmidrule(r){1-6}
    \multicolumn{2}{l}{PPO}       & $96.88\pm0.67\%$      & $46.67\pm 20.5\%$      & $72.75\pm3.99\%$       & $43.33 \pm17.00\%$      \\ 
    \multicolumn{2}{l}{PPO+DR}   & $89.96\pm7.89\%$      & $50 \pm 14.14\%$    & $ 73.29\pm2.32\%$      & $46.67 \pm 4.71\%$     \\ 
    \multicolumn{2}{l}{PPO+OODSI} &  $98.46\pm0.21\%$     & $66.67 \pm12.47\%$     & $75.67\pm8.07\%$       & $53.33\% \pm 9.43\%$     \\ 
    \multicolumn{2}{l}{PPO+DR+OODSI} &  $99.41\pm0.36\%$     & $\textbf{
    76.67} \pm4.71\%$     & $72.08\pm5.69\%$       & $\textbf{66.67} \pm 12.47\%$     \\ 
    \bottomrule
    \end{tabular}
\end{table*}

\begin{figure*}
    \centering
    \includegraphics[width=0.8\textwidth]{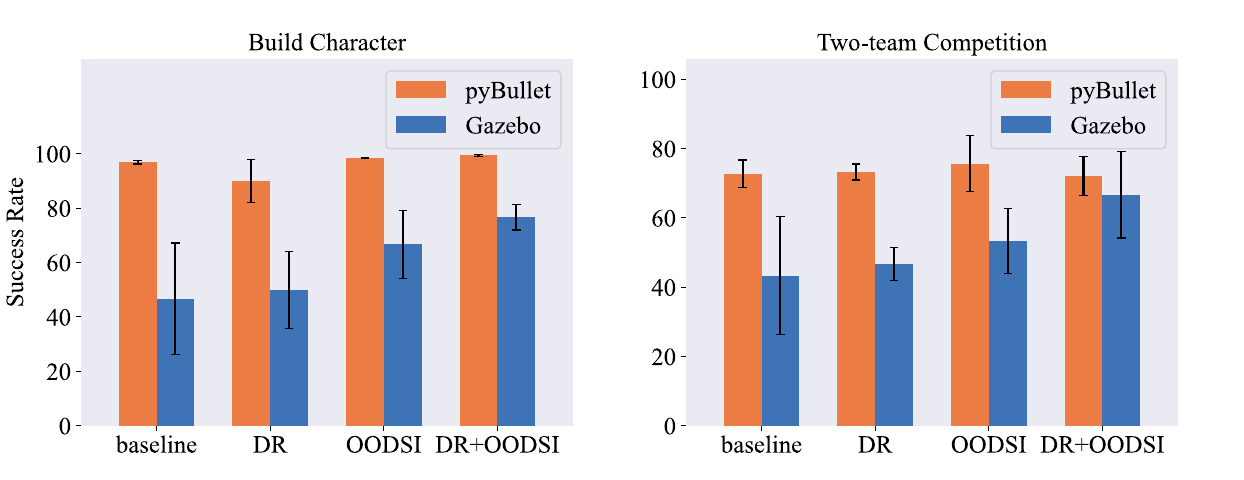}
    \caption{Sim2Real Performance Comparison}
    \label{fig:sim2real}
\end{figure*}

We train the agents to cooperate using the Centralized Training and Decentralized Execution (CTDE) framework~\cite{foerster2018counterfactual,lowe2017multi}. 
During training, we have a centralized value function, which processes the input information collected by all the agents and estimates the value i.e., estimated return, and assigns the credits to each agent in the team. 
During execution, each agent only needs to complete its own subtasks. 
\begin{figure}
    \centering
    \includegraphics[width=0.45\textwidth]{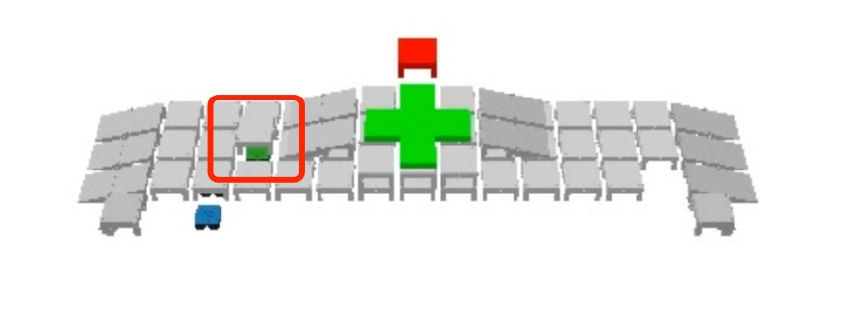}
    \caption{Blocking behavior in the pyBullet environment}
    \label{fig:two_floor_block_bullet}
\end{figure}
The agent's policy is represented by a multi-layer perceptron (MLP) network. 
To improve learning efficiency, the agents in the same team share the weights of the policy networks. 
We know that RL requires a large amount training samples to work well. 
To efficiently collect training samples, we adapt the large-scale distributed training framework in game AI~\cite{sun2020tleague,han2020tstarbot} and run the training process using Kubernetes.
For distributed training, we have two kinds of pods, i.e., the learner pods and the actor pods. 
The actor pods receive the policy weights from the learner pods and run rollouts in parallel to collect training samples. 
After collecting the samples, the actor pods send these samples to the learner pods. 
The leaner pods calculate the gradient using these samples and update the weights using Proximal Policy Optimization (PPO)~\cite{schulman2017proximal,yu2022surprising}. 
To avoid gradient explosion and stabilize the training process, we clip the gradient norm~\cite{mikolov2012statistical,pascanu2013difficulty}.
To train the two teams to compete with each other, after one team reaches a certain success rate threshold, we pause training for the leading team, fix its policies, and only train the other team. 
For the learning team, the paused team can be seen as a part of the environment. 
Since the policy of the paused team is fixed, the environment is stationary, which is suitable for learning.
If the environment is non-stationary, it is very difficult to train the agents to find optimal policies.
When the weaker team finds a counter policy and its success rate is higher than the threshold, we will switch the role for pause. 
This enables the two teams to stably evolve their policies during self-play.

\section{EXPERIMENTAL RESULTS}
\begin{figure}
    \centering
    \includegraphics[width=0.45\textwidth]{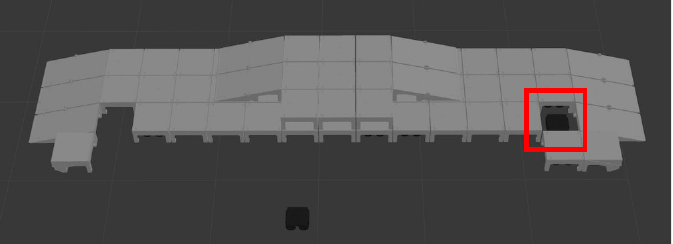}
    \caption{Blocking behavior in the Gazebo environment}
    \label{fig:two_floor_block_gazebo}
\end{figure}
\begin{figure}
    \centering
    \includegraphics[width=0.41\textwidth]{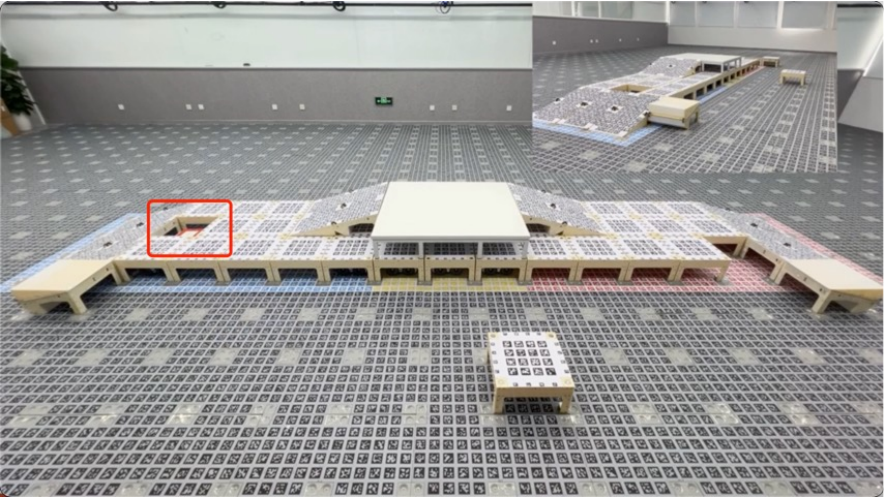}
    \caption{Blocking behavior in the real world}
    \label{fig:two_floor_block_real}
\end{figure}
We evaluate the proposed method in the robot craft arena, as introduced in Section~\ref{sec:robot_craft_arena}.
The robot craft arena consists of robot cars, blocks, and slopes. 
The robot cars can lift and move the blocks and slopes. 
Additionally, the robot car can also fold and unfold the slopes. 
We design two different kinds of tasks, including a collaboration task and a competition task. 
For the collaboration task, the robot cars cooperate with each other to build characters "TPX". 
For the competition task, the robot cars are divided into two teams. 
These two teams compete with each other to build two-floor structures with limited resources. 
We use pyBullet simulation for training and test the performance in the more realistic Gazebo simulation. 
To improve the training efficiency, we employ a distributed reinforcement learning training framework named TLeague~\cite{sun2020tleague}. In the experiments, we first conduct ablation studies on action masking. 
Secondly, we verify the effectiveness of the proposed OODSI method for Sim2Sim or Sim2Real transfer, compared to baseline methods, including PPO~\cite{schulman2017proximal} and Dynamics Randomization (DR)~\cite{peng2018sim}. 

\subsection{Action Masking Ablation Study}
To verify the effectiveness of guided RL via action masking, introduced in Section~\ref{sec:action_mask}, on improving training efficiency, we compare the performance of baseline method PPO with and without action masking in two different tasks. The agents in both of the tasks are trained for $3\times10^6$ steps. Figure~\ref{fig:action_mask_collaboration} and Figure~\ref{fig:action_mask_competition} show the learning curve of the normalized reward during training. The shadow area in the figures represents the variance. We can observe that guided RL via action masking can greatly improve the convergence speed and reduce the training time.

\subsection{Sim2Real Performance Comparison}

\begin{figure}
    \centering
    \includegraphics[width=0.40\textwidth]{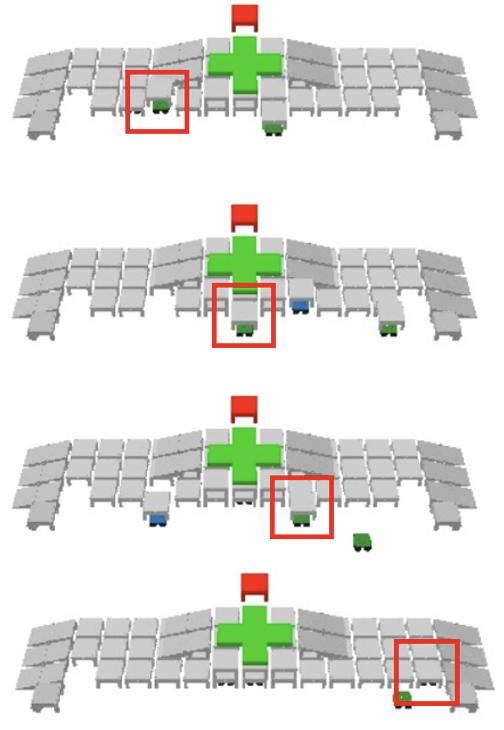}
    \caption{Stealing behaviour}
    \label{fig:steal}
\end{figure}

In this section, we conduct experiments to compare the Sim2Real performance of different methods in two different tasks, including collaboration and competition. 
We compared the success rates of four methods: PPO, PPO+DR, PPO+OODSI, and PPO+DR+OODSI in pyBullet and Gazebo environments. 
For all methods, we only train the models in the pyBullet environments and evaluate the Sim2Real performance in the Gazebo environments.
For DR, we randomly replace the actions using the stop action with 30\% probability to simulate action executing failure to improve robustness.
For OODSI method, we use the models trained for $3\times10^6$ steps to execute several episodes in the Gazebo environment to collect trajectories. Then, we sample $3$ trajectories and divide them equally into $5$ segments, collecting the first state of each segment to construct the initial state set for training. 
For each method, we selected models trained with $3$ different seeds and tested them in the Gazebo environment for $10$ episodes each. 
In the end, we calculated the success rate of each method over 30 episodes. 
The detailed results are provided in the table \ref{tab:sim2real}.
In Table~\ref{tab:sim2real} under Gazebo, we can see that, in both tasks, DR performs better than the baseline PPO. 
OODSI outperforms DR. The best performer is the combination of DR and OODSI. 
We also represent the experimental results in bar charts, see Figure~\ref{fig:sim2real}.
Figure~\ref{fig:sim2real} left part shows the success rate of the building character task. 
DR slightly improves the success rate in the Gazebo environment. 
OODSI method brings higher performance improvements. 
DR+OODSI significantly increases the success rate, which is $30\%$ higher compared to the baseline method.
Figure~\ref{fig:sim2real} right part shows the success rate of the two-team competition task.
In this task, the success rate is the sum of the winning rates of both teams. 
Similar to the results in the building character task, the DR and OODSI methods improve the Sim2Real performance. 
DR+OODSI still brings a significant increase, i.e., $23.34\%$, in the success rate. 
We can see that OODSI helps the agents learn robust multi-agent cooperative and competitive policies against the Sim2Real gap. 

\subsection{Qualitative Results}



In this section, we present the learned high-level strategies that we observed from the agent's policy. 
In the two-floor competition task, the robot car learns the blocking behavior. 
The robot car of the green team, which is highlighted with a red rectangle in Figure~\ref{fig:two_floor_block_bullet}, invades the blue team's territory and raises a block to prevent the left team from completing the task. 
Similarly, in both the Gazebo environment, see Figure~\ref{fig:two_floor_block_gazebo}, and the real-world robot craft arena, see Figure~\ref{fig:two_floor_block_real}, we also observe the blocking strategy. The robot car learns to go to the opponent's field and assist their own team to win the game by blocking the opponent team from completing the task.
Another interesting strategy the agent learns is 'stealing' the block, as shown in Figure~\ref{fig:steal}.
Figure~\ref{fig:steal} shows that a robot car of the green team 'steals' a block from the blue team's structure brings it to its own team and completes the building task.
The complete real-world process of building the character and the two-team competition tasks are shown in the video: \url{https://youtu.be/G_oLeJfXIyU}.

\section{CONCLUSIONS}
In this work, to push the boundary of Multi-Agent Reinforcement Learning (MARL) in the real world. 
We build a robot craft arena and propose collaborative and competitive tasks to train and evaluate the agent policies in different simulations and on the physical robots.  
We conduct ablation studies to show the effectiveness of guided RL via action masking in real-world robotic tasks.
To tackle the multi-agent Sim2Real challenge, we propose Out of Distribution State Initialization (OODSI) method and show its superior performance in combating the Sim2Real gap, compared to baseline methods.
We conduct multiple experiments in simulations and in the real world, demonstrating that the proposed method exhibits favorable Sim2Real performance and sheds light on practical applications of MARL in robotics.

\section*{APPENDIX}
We summarize the hyperparameters that we used in the experiments in Table~\ref{tab:hyperparameters}.
\begin{table}[htbp]
    \caption{Hyperparameters}
    \label{tab:hyperparameters}
    \renewcommand{\arraystretch}{1.1}
    \centering
    \begin{tabular}{p{0.24\textwidth}p{0.06\textwidth}}
    \hline
    \textbf{Parameter} & \textbf{Value} \\
    \hline
    Number of Layers & $4$ \qquad\\
    Number of Units & $256$ \qquad\\
    Activation Function & ReLU \qquad\\
    Learning Rate & $0.00001$ \qquad\\
    batch size & $4096$ \qquad\\
    Reward Discount & $0.95$ \qquad\\
    Optimizer & Adam \qquad\\
    GAE $\lambda$ & $0.95$ \qquad\\
    Replay Buffer Size & $256000$  \qquad\\
    Episode max step & 500 \qquad\\
    \hline
    \end{tabular}
\end{table}









\bibliographystyle{IEEEtran}
\bibliography{IEEEabrv}


\end{document}